\documentclass[11pt,letterpaper]{article}
\usepackage{ijcnlp2017}
\usepackage{times}
\usepackage{latexsym}
\usepackage{graphicx}
\usepackage{stmaryrd}
\usepackage{amsmath}
\usepackage{caption}
\usepackage[ruled]{algorithm}   
\usepackage[noend]{algpseudocode} 
\ijcnlpfinalcopy

\def\ijcnlppaperid{***}

\title{Fast BTG-Forest-Based Hierarchical Sub-sentential Alignment}

\author{Hao Wang \and Yves Lepage \\
  Graduate School of Information, Production and Systems,\\
  Waseda Univeristy\\
  {\tt \{oko\_ips@ruri., yves.lepage@\}waseda.jp}}

\date{}

\begin{document}

\maketitle

\begin{abstract}
In this paper, we propose a novel BTG-forest-based alignment method. Based on a fast unsupervised initialization of parameters using variational~IBM models, we synchronously parse parallel sentences top-down and align hierarchically under the constraint of BTG. Our two-step method can achieve the same run-time and comparable translation performance as {\tt{fast\_align}} while it yields smaller phrase tables. Final SMT results show that our method even outperforms in the experiment of distantly related languages, e.g., English--Japanese. 
\end{abstract}

\section{Introduction}
Bracketing transduction grammars (BTGs)~\cite{wu1997stochastic} are known to produce high quality, phrase-friendly alignments \cite{xiong2010learning,wang2007binarizing} for phrase-based statistical machine translation (SMT) \cite{koehn2003statistical} or syntax-based machine translation \cite{chiang2007hierarchical}. 

Differing from generative methods \cite{och2003systematic,liang2006alignment} that the complexity of word alignment grows exponentially with the length of the source and the target sentences, e.g., IBM models \cite{brown1993mathematics} and HMM-based model \cite{vogel1996hmm}, BTG provides a natural, polynomial-time, alternative method to reduce the search space in aligning. It also eliminates the need for any of the conventional heuristics.
 
Since BTG is effective to restrict the exploration of the possible permutations and alignments, there has been some interest in using BTGs for the purpose of alignment \cite{wu1995stochastic, zhang2005stochastic,wang2007binarizing,xiong2010learning,neubig2011unsupervised,neubig2012machine}. 
In particular, \newcite{cherry2007inversion} presented a phrasal BTG to the joint phrasal translation model and reported the results on word alignment. \newcite{haghighi2009better,riesa2010hierarchical} showed that BTG, which captures structural coherence between parallel sentences, helps in word alignment. \cite{saers2009learning} explored approximate BTG parsing and probabilistic induction for word alignment. \newcite{neubig2011unsupervised} incorporated a Gibbs sampling into joint phrase alignment and extraction framework. \newcite{kamigaitounsupervised}
modified the bidirectional agreement constraints and applied a more complex version (BTG-style
agreement) to train the BTG model jointly.

However, state-of-the-art BTG-based alignment methods are considered much time-consuming than the simplified generative model \cite{dyer2013simple}. The biggest barrier to applying BTG for alignment is the time complexity of na\"{\i}ve CYK parsing ($O(n^6)$), which makes it hard to deal with long sentences or large grammars in practice. Most of the previous research attempts to reduce the computational complexity of BTG parsing with some pruning methods. \newcite{zhang2005stochastic} propose \textit{tic-tac-toe} pruning by extending BTG with the additional lexical information based on IBM model~1 Viterbi probability. \newcite{haghighi2009better} investigate pruning based on the posterior predictions from two joint estimated models. \newcite{li2012beam} present a simple beam search algorithm for searching the Viterbi BTG alignments. 

In this paper, we propose a novel and fast BTG-parsing based word alignment method, which works as a heuristic to explore probable alignments in a given alignment matrix. It can be regarded as a hybridization of BTG parsing and IBM. We improved \cite{lardilleux2012hierarchical} with k-best beam search and introduced several new fast ways to build soft matrices using IBM models. Our aligner works as a top-down parser to generate hierarchical many-to-many symmetric alignments directly. We compare it with state-of-the-art methods and prove that it can lead to higher quality output for SMT. 

\section{From Viterbi Alignment to Bipartite Graph Bipartitioning}
Given a source sentence $F$ and a target sentence $E$, alignment associations between the bilingual sentences can be represented as a contingency matrix, which we note as $\mathcal{M}(F,E)$ \cite{matusov2004symmetric,liu2009weighted}. 

Given this adjacent matrix, there exist a number of methods to extract \textit{1-to-1} alignments or directly extract \textit{many-to-many} alignments from it. For example, \newcite{liu2010discriminative} propose a linear model to score the word alignments for searching the best one. These supervised approaches work using a large number of features \cite{haghighi2009better,liu2010discriminative}. We focus on simple unsupervised alignment. Other works are trying to induce BTGs with supervised \cite{haghighi2009better,burkett2010joint} or unsupervised \cite{wu1997stochastic,zhang2005stochastic} training, but they have a common disadvantage: they are time-consuming. In fact, BTG parsing is as simple as what we will discuss in the following. 

Consider a bipartite graph $G(U,V,\mathcal{E})$ with representing the matrix $\mathcal{M}$, with $\{U,V\}$ two independent subsets of vertices and $\mathcal{E}$ a set of edges. Each pair of nodes $(f,e)$ is connected with a weighted edge. With the constraints of BTGs, synchronously parsing a sentence pair $\langle F, E\rangle$ is a top-down processing that is equivalent to recursively bi-partitioning the graph $G$ into two disjoint sets of words $U$ and $V$ across languages.
For example, assume splitting the source sentence $F=\{X,\bar{X}\}$ (splitting at index $j$, between $f_j$ and $f_{j+1}$) and the target sentence $E=\{Y,\bar{Y}\}$ (splitting at index $i$, between $e_i$ and $e_{i+1}$) in a dichotomous way, i.e., \textit{straight} $\{U:XY,V:\bar{X}\bar{Y}\}$ or \textit{inverted} $\{U:X\bar{Y},V:\bar{X}Y\}$. Recursively bi-partitioning in $G$ will finally derive a BTG parse tree in which each leaf stands for a word-to-word correspondence. 
In this explanation, applying BTG parsing to a sentence pair can be regarded as trying to find the most reasonable splitting points $(i, j)$ in $F$ and $E$ at the same time. In the case of \textit{straight}, the optimal partition of such a graph is to find the minimum of the risk when reducing $G(U,V,\mathcal{E})$ to two subgraphs $G(X, Y, \mathcal{E}_{XY})$ and $G(\bar{X}, \bar{Y}, \mathcal{E}_{\bar{X}\bar{Y}})$ with applying BTG rule at $(i,j)$, at which the risk of reducing (cutting) $cut(U,V)$ (or $cut(i, j| \gamma)$) can be computed as total weight of the removed edges as:
\begin{equation}   
\resizebox{\hsize}{!}{$ 
cut(i, j| \gamma) = 
\begin{cases}
asso(X,\bar{Y}) +asso(\bar{X},Y), &  \gamma=straight\\ 
asso(X,Y) +asso(\bar{X},\bar{Y}), &  \gamma=inverted
\end{cases}   
$}
\end{equation} 
where 
\begin{equation}
\resizebox{0.6\hsize}{!}{$ 
asso(X, Y) = \sum_{f \in X}^{} \sum_{e \in Y}^{} w(f,e)  
$}
\end{equation} 
However, the minimum cut criterion favors cutting small sets of isolated nodes in the graph. To solve this problem, \newcite{shi2000normalized} propose a \textit{normalized cut} (\textit{Ncut}) to compute the cost as a fraction of the total edge connections to all the nodes in the graph. Following \cite{vilar2005experiments}, \newcite{lardilleux2012hierarchical} use \textit{Ncut} for sub-sentential alignment, with a na\"{i}ve assumption that words in a language are independent from each other as: 
\begin{align} 
\resizebox{0.22\hsize}{!}{$Ncut(i, j | \gamma)$}
&= 
\resizebox{0.55\hsize}{!}{$\frac{cut(i, j | \gamma)}{cut(i, j | \gamma) + 2\times cut_{left}(i, j | \bar{\gamma})}$} \nonumber \\
&+ 
\resizebox{0.55\hsize}{!}{$\frac{cut(i, j | \bar{\gamma})}{cut(i, j | \bar{\gamma})  + 2 \times cut_{right}(i, j | \bar{\gamma})}$}
\end{align}
$\bar{\gamma}$ is just the opposite of $\gamma$. The ideal criterion $\text {\it Ncut}$ for a recursive partitioning algorithm should minimize the disassociation between the unaligned blocks while maximizing the association within the aligned blocks at the same time.  
The time complexity of such a top-down algorithm is $O( m \times n \times \log \min(m,n))$, better than an exhaustive BTG bi-parsing algorithm which is known to be in $O(m^3 \times n^3)$.
\section{Forest-based BTG Alignment}
\newcite{lardilleux2012hierarchical} employs best-1 parsing to find the optimal \textit{Ncut}, which is intended to minimize. They binary segment the alignment matrix recursively to compute BTG-like alignments based on word level association scores but have not reported the alignment performance independently. While experimentally, we found that this strategy does not ensure the best global derivation. Different from that, we propose a BTG-forest-based parsing/alignment method with a beam search. Firstly, we define a scoring function $Score()$ aiming to find the best derivation $\tilde{\mathcal{D}}$ with the minimal value:
\begin{equation} 
\resizebox{0.56\hsize}{!}{$
\tilde{\mathcal{D}} =\underset{\mathcal{D}}{\mbox{argmin }} Score(\mathcal{D}_{Ncut}| \mathcal{M} )
$}
\end{equation}
$Ncut$ can be expressed as the arithmetic mean of two F-measures between $U$ and $V$. For example, in the \textit{straight} case, when $\{U: XY, V:\bar{X}\bar{Y}\}$:
\begin{equation}
\resizebox{0.9\hsize}{!}{$
F_{avg}(U, V) =  \frac{F_1 (X,Y) +F_1 (\bar{X},\bar{Y})}{2} =  1- \frac{Ncut(U, V)}{2} 
$ } 
\end{equation} 
With this expression, minimizing \textit{Ncut} is equivalent to maximizing $F_{avg}$. Intuitively, it suffices to replace \textit{Ncut} with $F_{avg}$ to derive the following formula, which gives the probability of a parsing tree, i.e., the probability of a sequence of derivation $\mathcal{D}$. The the best derivation $\tilde{\mathcal{D}}$ and the best word alignment $\mathbf{\hat{a}}$ can be defined as,
\begin{align} 
\resizebox{0.035\hsize}{!}{$\tilde{\mathcal{D}}$} 
&= 
\resizebox{0.44\hsize}{!}{$\underset{\mathcal{D}}{\mbox{ argmax }}  Score(\mathcal{D}_{f_{avg}}|\mathcal{M} )$}\\
&= 
\resizebox{0.44\hsize}{!}{$\underset{d_k \in \mathcal{D}}{\mbox{ argmax }} \prod_{k=1}^{K} {F_{avg}} (d_k)$} \\
\mathbf{\hat{a}}
&=
\resizebox{0.18\hsize}{!}{$Proj(\tilde{\mathcal{D}})$}  
\end{align}
Here, $d_k$ denote the operation of derivation at step $k$ during parsing, defined as a triple $\langle i, j, \gamma \rangle$, where $i,j$ are the splitting indices and $\gamma$ is either \textit{straight} or \textit{inverted}. Our incremental top-down BTG parsing algorithm with beam search is presented in \textbf{Algorithm}~\ref{alg:the_top_down_alg}. 
\begin{algorithm}[t]  
   \footnotesize
   \caption{\footnotesize Top-Down Parsing}
   \label{alg:the_top_down_alg}
    \begin{algorithmic}[1]
      \Function{ TopDownParsing}{$F$, $E$, $\tau$}
        \State $\mathcal{M} \leftarrow$ \Call{{InitializeSoftMatrix}}{$F$, $E$, $\tau$}
        \State $S_0 \leftarrow$ $\{$\Call{{InitializeState}}{0, $|F|$, 0, $|E|$}$\}$   
        \State $S_{final} \leftarrow \{ \} $  
        \For{$ l=0$ to $\mbox{min}(|F|, |E|)$} 
            \ForAll{s  $\in S_{l}$} 
              \ForAll{$s' \in$ \Call{{NextStates}}{$s$, $\mathcal{M}$}}
                  \State $S_{l+1} \leftarrow  S_{l+1} \cup s'$
                  \If {\Call{{IsTerminal}}{$s'$}} 
                     \State $S_{final} \leftarrow  S_{final} \cup s'$ 
                \EndIf
            \EndFor 
          \EndFor
            \State $S_{l+1} \leftarrow$ \Call{top}{$k$, $S_{l+1}$}
        \EndFor
        \State $\tilde{\mathcal{D}} = \underset{\mathcal{D}=\tilde{s}.\mathcal{D}, \tilde{s} \in S_{final}}{\mbox{ argmax }}  Score(\mathcal{D}|\mathcal{M} )$
        \State \Return $\tilde{\mathcal{D}}$
        \EndFunction
\end{algorithmic} 
\end{algorithm} 
We consider that the incremental parser has a parser state at each step. The state is defined as a four-tuple $\langle \mathcal{P}, \mathcal{D}, v, \tau\rangle$. $\mathcal{P}$ is the stack of unparsed blocks. $\mathcal{D}$ is the list of previous derivations $\{d_0,\dots,d_{l-1}\}$. A block denoted by $([i_0, i_1),[j_0, j_1))$ covers the source words from $f_{j_0}$ to $f_{j_1-1}$ and the target words from $e_{i_0}$ to $e_{i_1-1}$. $v$ records the current score. $\tau$ is set to \textit{true} on termination (stack $P$ is empty) and is \textit{false} elsewhere. At the beginning, the initial state contains only a block which covers all the words in $F$ and $E$. The block is split recursively and the the node type $\gamma$ (\textit{straight} or \textit{inverted}) is decided when the splitting point is determined according to the defined score function. $\Call{top}{k, S}$ returns the first $k$-th states from $S$ in terms of their scores $v$.  

The computational complexity of the top-down parsing algorithm is $O( k \times n  \times m \times \log \min(m,n))$ for sentence lengths $n$ and $m$, with a beam size of $k$. $\log \min(m,n)$ stands for the parsing depth. For each iteration, each state in the history will be used to generate new states as shown in \textbf{Algorithm}~\ref{alg:expand_hypo_alg}. \textbf{Algorithm}~\ref{alg:the_top_down_alg} terminates when no new hypothesis is generated or when it has reached the maximum number of iterations $\min(m,n)$.
\begin{algorithm}[t]
   \footnotesize
   \caption{Updating States}
   \label{alg:expand_hypo_alg}
    \begin{algorithmic}[1]
      \Function{NextStates}{$s, \mathcal{M}$}
      \State $S \leftarrow \{ \} $  
      \ForAll{$block \in s.P$} 
            \State $\{[i_0,i_1),[j_0,j_1)\} \leftarrow block$
            \State $ \mathcal{M}' \leftarrow \mathcal{M}_{[i_0,i_1),[j_0,j_1)}$
            \ForAll{$\{i, j\} \in \mathcal{M}'$}
              \For{$\gamma \in \{${\textit{straight,inverted}}$\}$}
              \State $d \leftarrow \langle i, j, \gamma\rangle$
                \State $v = s.v+$ \Call{$F_{avg}$}{$\mathcal{M}',d$}
                \State $block_{1,2}=$\Call{{SubMatrices}}{$\mathcal{M}', d$} 
                \State\Comment{{\textit{straight}: diagonal sub-matrices}} 
                \State\Comment{{\textit{inverted}: anti-diagonal sub-matrices}} 
                \State $s'=$\Call{{Update}}{$s,d,v,block_{1,2}$} 
                 \State $S \leftarrow  S \cup s' $
                 \EndFor
            \EndFor
        \EndFor 
    \State \Return $S$
    \EndFunction
\end{algorithmic}
\end{algorithm}

For the initialization of the matrix, there is a number of ways to define the weights of $w(f, e)$. The simplest one is to use the posterior probabilities of IBM model~1. \cite{moore2005association} pointed at several disadvantages of IBM model~1: it is either too sensitive to rare words or over-weights frequent words (like function words). For this reason, we incorporate variational Bayes (VB) into our model as proposed in \cite{riley2012improving}. We assume the distribution of the target vocabulary to be a Dirichlet distribution, with a symmetric Dirichlet prior as $\theta(f|e) \sim \mbox{Dirichlet}(\alpha)$\footnote{$\alpha=0.01$}. After computation of the posterior probabilities with the EM algorithm, the symmetrical score of $\theta(f, e)$ is defined as the geometric mean of the lexical translation probabilities in both directions $p(f|e)$ and $p(e|f)$.
\begin{align}
\resizebox{0.15\hsize}{!}{$w(f_j,e_i)$}
&= \resizebox{0.65\hsize}{!}{$ e^{\frac{\theta(f_j,e_i)}{\sigma_{\theta} }}\times 
	\begin{cases}
	 \quad\quad p_0 & \text{ otherwise}  \\ 
	 e^{\frac{\delta(i, j, n ,m)}{\sigma_{\delta} }}  & \text{ if }{h< r  }
	\end{cases} 
	$}\\ 
& 
\resizebox{0.64\hsize}{!}{$\theta(f_j,e_i)=\log(\sqrt{\theta(f|e) \times \theta(e|f)})$}\\ 
&
\resizebox{0.71\hsize}{!}{$\delta(j,i,n,m)=  \log(1-h(j,i,n,m))$} 
\end{align}  
where $\theta(f_j,e_i)$ is a word-to-word translation model and $\delta(j,i,n,m)$ is a distortion model. $r$ is a distortion threshold depends on language. $\sigma_\theta$ and $\sigma_\delta$ are hyper-parameters and $h(j,i, n ,m)=|j/n-i/m|$. Although this is not mandatory, we adjust values to a specified range $w(f_j,e_i) \in [{p_0}^2, 1), p_0=10^{-4}$. Since $\text {\it Ncut}$ is a normalized score, it does not require any normalization term. The hyper-parameters $\sigma_\theta$ and $\sigma_\delta$ are fixed at the beginning of experimentation by maximizing the \textit{Recall} in the preliminary experiments.
\section{Experiments} 
For evaluation of word alignment, we use the KFTT Corpus\footnote{http://www.phontron.com/kftt/} for English--Japanese. In the case of {\tt {GIZA++}} and {\tt {fast\_align}}, we train word alignments in both directions with the default settings, i.e., the standard bootstrap for IBM model~4 alignment in {\tt {GIZA++}} $(1^5 H^5 3^3 4^3)$ and $5$ iterations for {\tt {fast\_align}}. We then symmetrize the word alignments using \textit{grow-diag-final-and} (+\textit{gdfa}) and evaluate with the final obtained alignments. Perhaps some comparison with other BTG alignment methods is necessary to confirm the advantages of our proposed method. For this consideration, we use an open-sourced BTG-based word aligner, {\tt {pialign}}\footnote{http://www.phontron.com/pialign/}. We run it with 8 threads and train the model with batch size 40 and only taking 1 sample during parameter inference. We extract phrases directly from the word-to-word alignment (many-to-many) with traditional heuristic \cite{koehn2003statistical} for translation.
For our implementation, named {\tt {Hieralign}}, we limit the run-time to that of {\tt {fast\_align}} for fairness. We perform 5 iterations EM estimation using IBM~1 with variational Bayes, with a beam size of 10 during parsing. Since reestimation of the Viterbi probability with the \textit{gdfa} heuristic (+VBH) is very fast, we also employ it before the step of the parsing.
\begin{table}
\captionsetup{font=small}
\centering
\resizebox{0.45\textwidth}{!}{
\begin{tabular}{|l|r|r|r|}
\hline
                     & {\bf AER }   &  {\bf Time}    & {\bf Size (M)}                      \\ \hline
{\tt {fast\_align }}         & 52.08   & 1:40    & 7.40                \\\hline
{\tt {GIZA++ }}              & 42.39   & 51.38     & 7.98 \\ \hline
{\tt {pialign}} (many-to-many)      & 57.11 & 199:46    & 7.73                        \\ \hline
{\tt {Hieralign}}($\sigma_\theta =1 $)  & 58.12  & 2:03 & 4.91                         \\
{\tt {Hieralign}} ($\sigma_\theta =3, \sigma_\delta =5$) & 53.99 & 2:20 &    5.50 \\ \hline

\end{tabular} }
\caption{Alignment error rate (AER), wall-clock time (minutes:seconds) required to obtain the symmetric alignments and phrase table size (\# of entries) on KFTT corpus.}\label{alignmenttable}
\end{table} 
For the phrase-based SMT task, we conduct experiments in English--German (en--de) using the WMT 2008 Shared Task\footnote{http://www.statmt.org/wmt08/shared-task.html}; English--Japanese (en--ja) using the KFTT corpus. For translation evaluation, training, development, test sets are independent. 

Table~\ref{alignmenttable} shows that our proposed method achieves competitive performance on the KFTT Corpus with state-of-the-art alignment methods. AER ({\tt {Hieralign}}) is behind fast align, even more
than GIZA++. However, \cite{lopez2006word,fraser2007measuring} question the link between this word alignment quality metrics and translation results. There is no proof that improvements in alignment quality metrics lead to improvements in phrase-based SMT performance. Since our method forces each source and target word aligned (\textit{many-to-many}), it is prone to generate fewer entries in the translation tables. We thus measured the sizes of the translation tables obtained. Phrase tables extracted from the alignments by {\tt {Hieralign}} are smaller by a third in comparison to those of the baseline. 

The accuracy of the translations produced by our method are compared to those produced by {\tt {GIZA++}} (+\textit{gdfa}), {\tt {fast\_align}} (+\textit{gdfa}) and {\tt {pialign}} in Table~\ref{mt_result}, in which standard automatic evaluation metrics are used: BLEU \cite{papineni2002bleu} and RIBES \cite{isozaki2010automatic}. There is no significant difference on the final results in en--de and even better in en--ja. Given the results in Table~\ref{mt_result} with the distortion feature ($\sigma_\theta =3, \sigma_\delta =5$) and without distortion feature ($\sigma_\theta =1 $), we can also draw the conclusion that adding the distortion feature slightly improves the alignment results. 

\begin{table}
\captionsetup{font=small}
\centering
\resizebox{0.48\textwidth}{!}{
\begin{tabular}{|l|ll|ll|}
\hline
&\multicolumn{2}{|c|}{\bf en-de} &\multicolumn{2}{|c|}{\bf en-ja} \\\hline
& {\bf BLEU} & {\bf RIBES} & {\bf BLEU} & {\bf RIBES}\\\hline
{\tt {fast\_align}} & 19.61 & 70.02 & 21.32& 68.10 \\\hline
{\tt {GIZA++}}      & \textbf{19.82} & \textbf{70.48} & 22.57$^\dag$&68.79 \\\hline
{\tt {pialign}} (many-to-many) & 19.81 & 69.99 & 21.69& \textbf{69.07} \\\hline
{\tt {Hieralign}} ($\sigma_\theta =1 $)  & 19.33 & 69.62 & \textbf{22.69}$^\dag$ & 67.94 \\
{\tt {Hieralign}} ($\sigma_\theta =3, \sigma_\delta =5$) & 19.55 & 69.90 & 22.57$^\dag$ & 68.35  \\  \hline
\end{tabular}  } 
\caption{BLEU and RIBES scores in phrase-based SMT experiments, $\dag$ means significantly different with {\tt {fast\_align}} baseline ($p<0.05$) \cite{koehn2004statistical}.}\label{mt_result}
\end{table}

\section{Conclusion}

To summarize, we proposed a novel BTG-forest-based top-down parsing method for word alignment, we improved \cite{lardilleux2012hierarchical} with better parameter initialization method and return a open-sourced software {\tt {Hieralign}}. We achieved comparable translation scores with state-of-the-art methods, while the speed is fast. For future work, we believe that incorporating neural models to build the soft-matrix for our method should make a positive influence.


\bibliography{ijcnlp2017}
\bibliographystyle{ijcnlp2017}

\end{document}